\documentclass[letterpaper]{article} 
\usepackage{aaai24}  
\usepackage{times}  
\usepackage{helvet}  
\usepackage{courier}  
\usepackage[hyphens]{url}  
\usepackage{graphicx} 
\usepackage{natbib}  
\usepackage{caption} 
\usepackage{algorithm}
\usepackage{algorithmic}
\usepackage{streams}

\usepackage{newfloat}
\usepackage{listings}
\DeclareCaptionStyle{ruled}{labelfont=normalfont,labelsep=colon,strut=off} 
\floatstyle{ruled}
\newfloat{listing}{tb}{lst}{}
\floatname{listing}{Listing}
\usepackage{amsmath}
\usepackage{amssymb}
\usepackage{amsthm} 
    \theoremstyle{plain}
    \newtheorem{theorem}{Theorem}[section]
    \theoremstyle{definition}
    \newtheorem{definition}[theorem]{Definition}
    \newtheorem{example}[theorem]{Example}
\usepackage{cleveref} 
\usepackage{mathtools} 

\usepackage{stringdiagrams} 

\newcommand{\mc}{\mathcal}

\title{Categorical Foundation of Explainable AI: A Unifying Theory}

\author {
    Pietro Barbiero$^*$\textsuperscript{\rm 1},
    Stefano Fioravanti$^*$\textsuperscript{\rm 2},
    Francesco Giannini$^*$\textsuperscript{\rm 2},\\
    Alberto Tonda\textsuperscript{\rm 3},
    Pietro Li\'o\textsuperscript{\rm 1},
    Elena Di Lavore\textsuperscript{\rm 4}
}
\affiliations {
    \textsuperscript{\rm 1}University of Cambridge, UK\\
    \textsuperscript{\rm 2}Universit\`a di Siena, Italy\\
    \textsuperscript{\rm 3}INRA, Université Paris-Saclay, France\\
    \textsuperscript{\rm 4}Tallinn University of Technology, Estonia\\
    \texttt{pb737@cam.ac.uk}, \texttt{stefano.fioravanti@unisi.it}, \texttt{francesco.giannini@unisi.it}
}

\begin{document}

\maketitle

\begin{abstract}
Explainable AI (XAI) aims to address the human need for safe and reliable AI systems. However, numerous surveys emphasize the absence of a sound mathematical formalization of key XAI notions---remarkably including the term ``\textit{explanation}'' which still lacks a precise definition. To bridge this gap, this paper presents the first mathematically rigorous definitions of key XAI notions and processes, using the well-funded formalism of Category theory. We show that our categorical framework allows to: (i) model existing learning schemes and architectures, (ii) formally define the term ``explanation'', (iii) establish a theoretical basis for XAI taxonomies, and (iv) analyze commonly overlooked aspects of explaining methods. As a consequence, our categorical framework promotes the ethical and secure deployment of AI technologies as it represents a significant step towards a sound theoretical foundation of explainable AI.
\end{abstract}

\section{Introduction}
Explainable AI (XAI) has emerged as a critical research field to address ethical~\citep{duran2021afraid,lo2020ethical} and legal~\citep{wachter2017counterfactual, gdpr2017} concerns related to the safe deployment of AI technologies. However, several domain surveys remark that key fundamental XAI notions still lack a formal definition, and that the whole field is still missing a unifying and sound formalism~\citep{adadi2018peeking,palacio2021xai}.

The notion of \textit{explanation} itself
still lacks a proper mathematical formalization, as demonstrated by the attempts to define the concept in the current literature: {\small\textit{``An explanation is an answer to a `}why?\textit{' question''~\cite{miller2019explanation}} or \textit{``An explanation is additional meta information, generated by an external algorithm or by the machine learning model itself, to describe the feature importance or relevance of an input instance towards a particular output classification''~\cite{Das2020OpportunitiesAC}}}.

We argue that the absence of a rigorous formalization of key explainable AI notions may significantly undermine progress in the field by leading to ill-posed questions, annoying clutter in taxonomies, and narrow perspectives on future research directions. In contrast, a sound mathematical definition would provide a solid foundation for XAI notions and taxonomies, leading to a more organized and efficient research field. This way, researchers could easily identify knowledge gaps and promising research directions---including the exploration of the (currently overlooked) theoretical side of explainable AI. 

\textbf{Contributions.} We propose a theoretical framework allowing a unified, comprehensive and rigorous formalization of foundational XAI notions and processes (\Cref{sec:framework}). To the best of the authors' knowledge, this is the first work investigating this research direction in the XAI field. To this objective, we use Category theory as it represents a sound mathematical formalism centered around \emph{processes}, and for this reason, it is widely used in theoretical computer science~\citep{Abramsky2004ACS, Selinger2001ControlCA, stein2021, Swan2022, Turi1997TowardsAM}, and, more recently, in AI~\citep{aguinaldo2021graphical,cruttwell2022categorical, ong2022learnable, shiebler2021category, katsumata19}. In particular, we show that our categorical framework enables us to: model existing learning schemes and architectures (\Cref{sec:finding1}), formally define the term ``explanation'' (\Cref{sec:syExp}),  establish a theoretical basis for XAI taxonomies (\Cref{sec:tax}), and analyze commonly overlooked aspects of explaining methods (\Cref{sec:xai-semantics}).

\section{Explainable AI Theory: Requirements}
\label{sec:pre}

In order to build a sound theory, we first need to identify (i) a set of relevant notions, objects, and processes, and (ii) a proper language to formalize them. To formalize objects---such as explanations---we rely on Institution Theory~\citep{goguen1992institutions} as it provides an abstract framework for formal languages' syntax and semantics (Sec. \ref{sec:inst}). To formalize explainable AI algorithms and their dynamics instead, we rely on Category Theory~\citep{eilenberg1945general} as it provides a mathematical framework specifically designed to analyze processes and their dynamics (Sec. \ref{sec:moncat}).

\subsection{Institution Theory: A Framework for Explanations}
\label{sec:inst}

To provide a formal definition of ``explanation'' (\Cref{sec:syExp}) we rely on institution theory~\citep{goguen1992institutions}.  
Institution theory offers an ideal platform for formalizing explanations---whether expressed through symbolic languages or semantic-based models---as it enables a thorough analysis of both the structure (syntax) and meaning (semantics) of explanations across diverse languages~\citep{tarski1944semantic}, thus facilitating a deeper understanding of their nature.

More rigorously, an institution $I$ consists of (i) a category \(\cat{Sign}_I\) whose objects are signatures (i.e. vocabularies of symbols), (ii) a functor $Sen: \cat{Sign}_I \mapsto \cat{Set}$ which provides sets of well-formed expressions ($\Sigma$-sentences) for each signature \(\Sigma\in\cat{Sign}_I^o\), and (iii) a functor $Mod: \cat{Sign}_I^{op} \mapsto \cat{Set}$\footnote{Given a category \(\cat{C}\), $\cat{C}^{op}$ denotes its \emph{opposite} category, formed by reversing its morphisms~\citep{maclane78}.} that assigns a semantic interpretation (i.e. a world) to the symbols in each signature~\citep{goguen2005concept}.

A typical example of institution is First-Order Logic (FOL), where the category of signatures is given by sets of relations
as objects and arity-preserving functions as morphisms. Sentences and models are defined by standard FOL formulas and structures.


\subsection{Category Theory: A Framework for XAI Processes}
\label{sec:moncat}

(X)AI processes all share three basic properties: (i) they map (multiple) inputs to (multiple) outputs via a composition of parametric operations (see \textit{monoidal categories}), (ii) they update the parameters of such operations based on some error function (see \textit{feedback categories}), and (iii) they keep updating such parameters over time until convergence (see \textit{cartesian streams}). In the following paragraphs we formalize these (X)AI properties using a categorical language.

\paragraph{A primer on categories: }
Intuitively, a category is a collection of objects and morphisms satisfying specific composition rules. 
\begin{definition}[\citet{eilenberg1945general}]
    A \emph{category} \(\cat{C}\) consists of a class of \emph{objects} $\cat{C}^o$ and, for every \(X,Y \in \cat{C}^o\), a set of \emph{morphisms} $\hom(X,Y)$ with input type \(X\) and output type \(Y\). A morphism \(f \in \hom(X,Y)\) is written \(f \colon X \to Y\), and for all morphisms \(f \colon X \to Y\) and \(g \colon Y \to Z\) there is a \emph{composite} morphisms \(f \dcomp g \colon X \to Z\), with composition being associative. For each \(X \in \cat{C}^o\) there is an \emph{identity} morphism \(\id{X} \in \hom(X,X)\) that makes composition unital.
\end{definition}

Well-known examples are the categories $\cat{Set}$, whose objects are \emph{sets} and morphisms are \emph{functions}, and $\cat{Vec}$, whose objects are \emph{vector spaces} and morphisms are \emph{linear maps}.
Different categories can be connected using special operators called \emph{functors} i.e., mappings of objects and morphisms from one category to another (preserving compositions and identity morphisms).  
For instance, there is a functor $\mc{F}$ from $\cat{Vec}$ to $\cat{Set}$ that simply ignores the vector space structure. 

\paragraph{Monoidal Categories: compose multi-input/output processes}
In this work we are mainly interested in \emph{monoidal categories} as they offer a sound formalism for processes with multiple inputs and outputs~\citep{Coecke2017,fritz2020}. Monoidal categories~\citep{maclane78} are categories with additional structure, namely a monoidal product $\times$ and a neutral element, enabling the composition of morphisms in parallel (cf. Appendix \ref{app:mon-cat}). Notably, monoidal categories allow for a graphical representation of processes using \emph{string diagrams}~\citep{joyal1991geometry}. String diagrams enable a more intuitive reasoning over equational theories, and we will use them throughout the paper to provide illustrative, yet formal, definitions of XAI notions. The Coherence Theorem for monoidal categories~\citep{maclane78} guarantees that string diagrams are a sound and complete syntax for monoidal categories. Thus all coherence equations for monoidal categories correspond to continuous deformations of string diagrams. For instance, given \(f \colon X \to Y\) and \(g \colon Y \to Z\), the morphisms \(f \dcomp g \colon X \to Z\) and \(\id{X}\) are represented as
\includegraphics[width=0.35\textwidth]{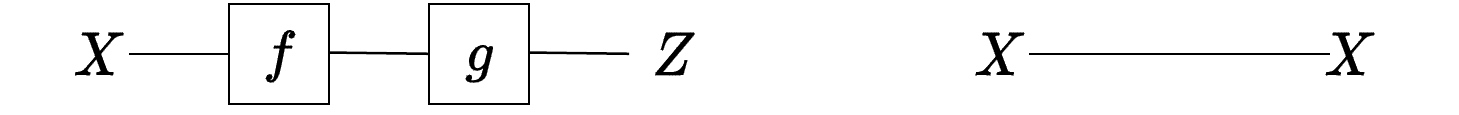};
the equation \(f \dcomp \id{Y} = f = \id{X} \dcomp f\) as

\includegraphics[width=0.4\textwidth]{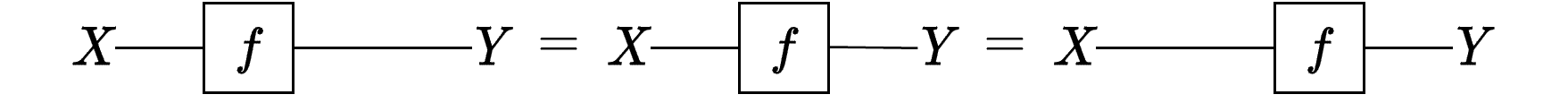};

the morphism $h$ with multiple inputs $X_1,X_2,X_3$ and outputs $Y_1,Y_2$ (left), and the parallel composition of two morphisms \(f_1 \colon X_1 \to Y_1\) and \(f_2 \colon X_2 \to Y_2\) (right) can be represented as follows:

\begin{figure}[!h]
    \centering\includegraphics[width=0.4\textwidth]{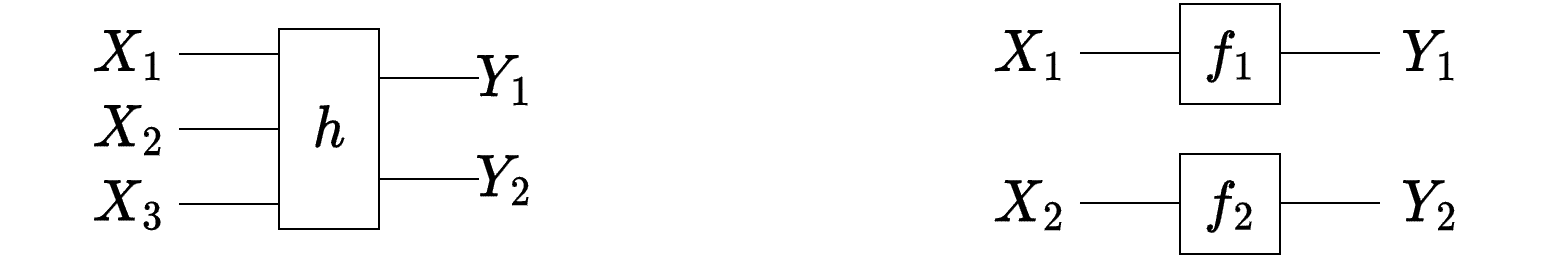}
\end{figure}

\paragraph{Feedback Categories: update process states}

A common process in machine learning involves the update of the parameters of a function, based on the \emph{feedback} of a loss function. To model this process, we can use \emph{feedback monoidal categories}.

\begin{definition} [\citep{katis02,di2021canonical}]
A \emph{feedback monoidal category} is a symmetric (cf. Appendix~\ref{app:symm-mon-cat}) monoidal category \(\cat{C}\) endowed with a functor $\mc{F}:\cat{C} \rightarrow \cat{C}$, and an operation \(\fbk[S] \colon \hom (X \times \mc{F}(S), Y \times S) \to \hom (X,Y)\) for all \(X,Y,S\) in \(\cat{C}^o\), which satisfies a set of axioms (cf.  \Cref{app:feedback-cat}).
\end{definition}

\paragraph{Cartesian Streams: dynamic update of processes over time}
In learning processes, optimizing the loss function often involves a sequence of feedback iterations. 
Following~\citet{katsumata19}, we use \emph{Cartesian streams} (cf. \Cref{app:streams}) to model this kind of processes. 
Cartesian streams form a feedback \emph{Cartesian} monoidal category, i.e are equipped, for every object \(X\), with morphisms \(\cp_X \colon X \to X \times X\) and \(\discard_X \colon X \to e\) that make it possible to copy and discard objects (cf. Appendix~\ref{app:symm-mon-cat}). This makes them the ideal category to formalize (possibly infinite) streams of objects and morphisms.

\begin{definition}[\citep{katsumata19,uustalu2008comonadic}] 
Let $\cat{C}$ be a Cartesian category. We call $\Stream{\cat{C}}$ the category of \emph{Cartesian streams} over $\cat{C}$, whose objects \(\stream{X} = (X_0, X_1, \dots)\) are countable lists of objects in $\cat{C}$, and given \(\stream{X},\stream{Y}\in \Stream{\cat{C}}^o\), the set of \emph{morphisms} $\hom(\stream{X}, \stream{Y})$ is the set of all \(\stream{f} \colon \stream{X} \to \stream{Y}\), where \(\stream{f}\) is a family of morphisms in $\cat{C}$, \(f_n \colon X_0 \times \cdots \times X_n \to Y_n\), 
for $n\in\mathbb{N}$.
\end{definition}

In Cartesian streams, a morphism \(f_n\) represents a process that receives a new input \(X_n\) and produces an output \(Y_n\) at time step \(n\). We can compute the outputs until time \(n\) by combining \(f_0, \dots, f_n\) to get \(\tilde{f}_n \colon X_0 \times \cdots \times X_n \to Y_0 \times \cdots \times Y_n\) as follows:
\vspace{-0.3cm}
\includegraphics[width=0.45\textwidth]{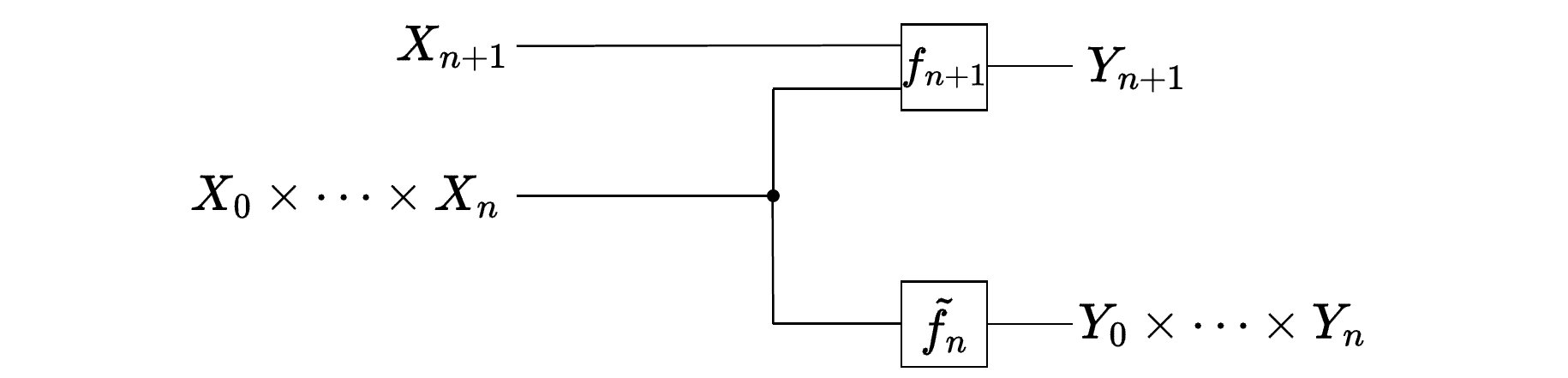}
\begin{itemize}
\vspace{-0.3cm}
\item \(\tilde{f}_0 \defn f_0\)
\item \(\tilde{f}_{n+1} \defn (\id{X_{n+1}} \times \cp_{X_0 \times \cdots \times X_n}) \dcomp (f_{n+1} \times \tilde{f}_n)\)
\end{itemize}

We denote by $X^{\mathbb{N}}$ the object $\stream{X}\in\Stream{\cat{C}}^o$ such that $\stream{X}=(X,X,\ldots)$, for some $X\in \cat{C}^o$.
Notably, Cartesian streams form a feedback monoidal category \citep{monoidalStreams} and thus are capable to 
model dynamic processes with feedback---such as learning processes, as we will show in~\Cref{ex:mlp} and~\ref{ex:nas}. 

\section{Categorical Framework of Explainable AI}
\label{sec:framework}

We use feedback monoidal categories and institutions to formalize fundamental (X)AI notions. To this end, we first introduce the definition of ``abstract learning agent'' (Section~\ref{sec:learning-agent}) as a morphism in free feedback monoidal categories, and then we describe a functor instantiating this concept in the concrete feedback monoidal category of \(\Stream{\Set}\) (Section~\ref{sec:interpretation}). Intuitively, a \emph{free} category serves as a template for a  class of categories (e.g., feedback monoidals). To generate a free category, we just need to specify a set of objects and morphisms generators. Then we can realize ``concrete'' instances of a free category $\cat{F}$ through a functor from $\cat{F}$ to another category $\cat{C}$ that preserves the axioms of $\cat{F}$ (cf. \Cref{app:free}). 

\subsection{Abstract Learning Agents}
\label{sec:learning-agent}

We formalize the abstract concept of an explaining learning agent drawing inspiration from ~\citep{cruttwell2022categorical, katsumata19, wilson2021reverse}. At a high level, learning can be characterized as an \emph{iterative process with feedback}. This process involves a function (known as \emph{model} or \emph{explainer} in a XAI method) which updates its internal states (e.g., a set of \emph{parameters}) guided by some feedback from the environment (often managed by an \emph{optimizer}). Formally, we define an abstract learning agent as a morphism in the free feedback Cartesian monoidal category $\mathsf{XLearn}$ generated by: 

\begin{itemize}
    \item the objects $X,Y,Y^*, P$, and $E$ representing input, output, supervision, parameter, and explanation types;
    \item the \textit{model/explainer} morphism $\eta: X \times P \rightarrow Y \times E$ which produces predictions in $Y$ and explanations in $E$;
    \item the \textit{optimizer} morphism \(\nabla_Y \colon Y^* \times Y \times P \to P\) producing updated parameters in \(P\) given supervisions in \(Y^*\), model/explainer predictions in \(Y\), and parameters in \(P\).
\end{itemize}

\begin{definition}
$\mathsf{XLearn}$ is the free feedback Cartesian category generated by the objects $X,Y,Y^*, P, E$ and by the morphisms $\eta: X \times P \rightarrow Y\times E$ and $\nabla_{Y}: Y^* \times Y \times P \rightarrow P$.
\end{definition}

The introduction of these morphisms allows us to establish a formal definition of an abstract learning agent.

\begin{definition} 
An \emph{abstract learning agent} is the morphism in $\mathsf{XLearn}$ given by the morphisms' composition:  $\fbk[P]\left((\id{Y^*\times X}\times\nu_{P});(\id{Y^*}\times \eta \times \id{P}); (\id{Y^*\times Y}\times\discard_{E}) ;\nabla_{Y}\right)$
\begin{center}\includegraphics[scale=0.70]{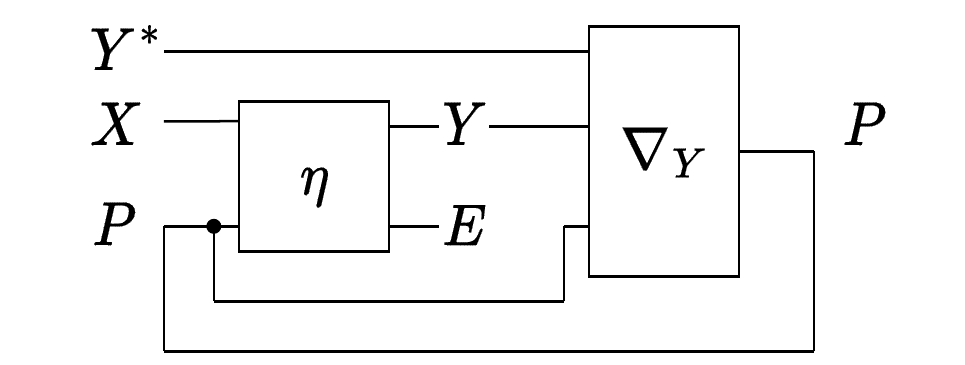}\end{center}
\end{definition}

\subsection{Concrete Learning and Explaining Agents}
\label{sec:interpretation}

The free category $\mathsf{XLearn}$ allows us to highlight the key features of learning agents from an abstract perspective. However, we can instantiate explaining learning agents in ``concrete'' forms using a feedback functor from the free category $\mathsf{XLearn}$ to the category of Cartesian streams over \(\Set\), i.e. \(\Stream{\Set}\). This functor can establish a mapping from our abstract construction to any concrete setting, involving diverse explainers (e.g., decision trees, logistic regression), input data types (e.g., images, text), supervisions, outputs, parameters, or explanations. Achieving this mapping requires the definition of a specific functor we call \emph{translator}.

\begin{definition} 
An \emph{agent translator} is a  feedback Cartesian functor  $\mc{T}: \mathsf{XLearn} \rightarrow \Stream{\Set}$.
\end{definition}

Among translators, we distinguish two significant classes: those that instantiate learning agents and those that instantiate explainable learning agents (Definitions \ref{def:concrete-la} and \ref{def:xla} respectively). Intuitively, a concrete learning agent is an instance of an abstract learning agent that does not provide any explanation while a concrete explaining learning agent does output an (non-empty) explanation.

\begin{definition}\label{def:concrete-la} 
Given an agent translator $\mc{T}$ with $\mc{T}(E) = \{*\}^{\mathbb{N}}$, where $\{*\}$ is a singleton set. A \textit{learning agent (LA)} is the image $\mc{T}(\alpha)$, being $\alpha$ the abstract learning agent.
\end{definition}
The set $\{*\}^{\mathbb{N}}$ denotes the neutral element of the  monoidal product in $\Stream{\Set}$ and conveys the absence of explanations. In this case, $\mc{T}(\eta)$ will be simply called \textit{model}, and we will remove the explicit dependence on $\{*\}$ in the output space as $T(Y)\times\{*\}^{\mathbb{N}}\iso T(Y)$. To instantiate explaining learning agents instead, we introduce two distinct types of translators: the semantic and the syntactic translator. This choice is motivated by the foundational elements of institution theory, namely sentences and models: Sentences correspond to well-formed \emph{syntactic} expressions, while models capture the \emph{semantic} interpretations of these sentences~\citep{goguen1992institutions}. We refer to concrete instances of both syntactic and semantic explaining learning agents as ``explaining learning agents'' (XLA).

\begin{definition}\label{def:xla} 
Let $\mc{T}$ be an agent translator, $I$ an institution and $\alpha$ the abstract learning agent. The image $\mc{T}(\alpha)$ is said a \textit{syntactic explaining learning agent} if $\mc{T}(E) = Sen(\Sigma)^{\mathbb{N}}$ and a \textit{semantic explaining learning agent} if $\mc{T}(E) = Mod(\Sigma)^{\mathbb{N}}$, for some signature $\Sigma$ of $I$.
\end{definition}

The high degree of generality in this formalization enables the definition of any real-world learning setting and learning agent (to the author knowledge). Indeed, using Cartesian streams as the co-domain of translators, we can effectively model a wide range of learning use cases, including (but not limited to) those involving iterative feedback. To simplify the notation, in the following sections we use the shortcuts: $\mathcal{X} = \mc{T}(X)$, $\mathcal{Y} = \mc{T}(Y)$,
$\mathcal{Y}^* = \mc{T}(Y^*)$,
$\mathcal{P} = \mc{T}(P)$, $\mathcal{E} = \mc{T}(E)$, $\hat{\eta}=\mc{T}(\eta)$, $\hat{\nabla}_{\mathcal{Y}}=\mc{T}(\nabla_Y)$, and $\mathcal{Y} = \mathcal{Y}^*$ when not specified. 

\section{Impact on XAI and Key Findings}

\subsection{Finding \#1: Our framework models existing learning schemes and architectures}
\label{sec:finding1}

As a proof of concept, in the following examples we show how the proposed categorical framework makes it possible to capture the structure of some popular learning algorithms such as neural networks.

\begin{figure*}[!ht]
    \centering
    \includegraphics[scale=0.6]{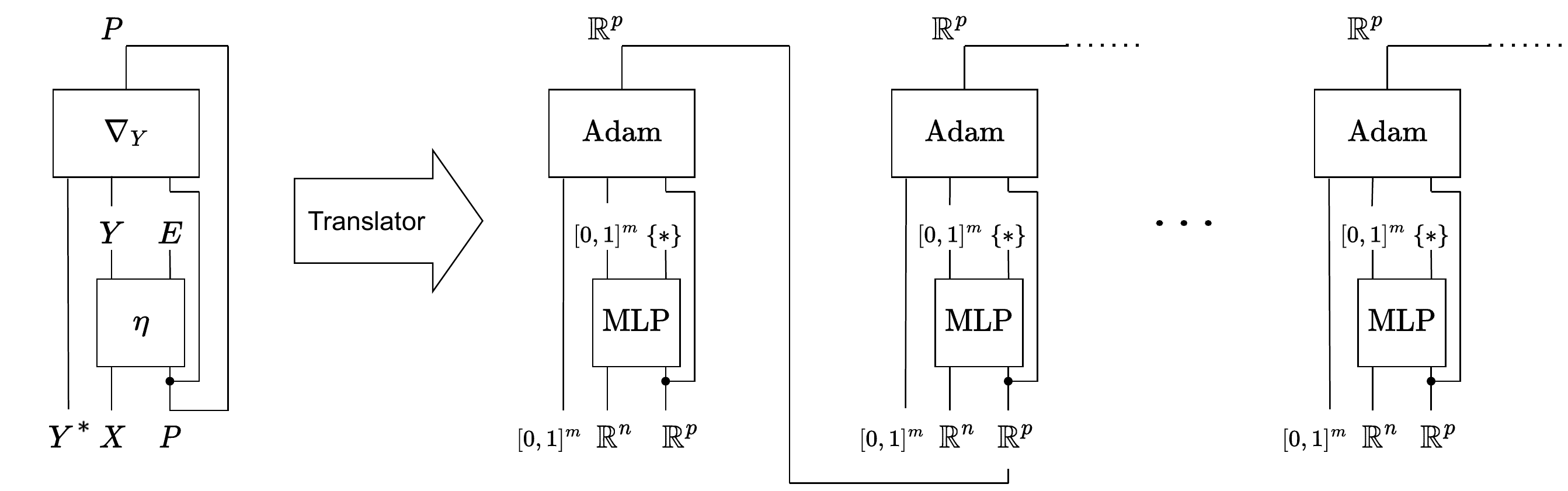}
    \label{fig:LA}
\end{figure*}

\begin{example}\label{ex:mlp} 
A classic multi-layer perceptron (MLP,~\citep{rumelhart1985learning}) classifier with Adam optimizer~\citep{kingma2014adam} is an instance of an abstract learning agent whose translator functor is defined as follows (\Cref{fig:LA}):
    $\mc{X}= (\mathbb{R}^n)^{\mathbb{N}}$, $\mc{Y} = \mc{Y}^* =\left([0,1]^m\right)^{\mathbb{N}}$, $\hat{\eta}_i$ being an MLP for all $i$, $\mc{P}$ the space of the MLP parameters, e.g. $\mc{P}=(\mathbb{R}^p)^{\mathbb{N}}$, $\hat{\nabla}_{\mc{Y}_i}$ being e.g. the Adam optimizer, and $\mc{E} = \{*\}^{\mathbb{N}}$.
\end{example}

In Example \ref{ex:mlp} we successfully model an MLP by constraining the components of the morphism $\eta$ to have constant values $\hat{\eta}_i = \text{MLP}$ (independent of the first $i-1$ inputs). However, by removing this constraint, we can instantiate a broader class of learning agents including e.g. recurrent neural networks~\citep{hochreiter1997long, hopfield1982neural} and transformers~\citep{vaswani2017attention}. In these scenarios, the neural functions become dependent on previous inputs, effectively capturing the input stream. Additionally, we can also model learning settings where a model's architecture changes over time, as in neural architecture search~\citep{elsken2019neural}.

\begin{example}\label{ex:nas} 
A classical Neural Architecture Search algorithm~\citep{elsken2019neural} is an instance of an abstract learning agent whose translator functor is defined as follows:
    $\mc{X}= (\mathbb{R}^n)^{\mathbb{N}}$, $\mc{Y}= \mc{Y}^* =([0,1]^m)^{\mathbb{N}}$, $\hat{\eta}_i = \text{MLP}_i$ being a different neural architecture for every step, $\mc{P}$ the space of the MLPs parameters, e.g. $\mc{P}=(\mathbb{R}^p)^{\mathbb{N}}$, $\hat{\nabla}_{\mc{Y}_i}$ being the Adam optimizer, and $\mc{E} = \{*\}^{\mathbb{N}}$.
\end{example}

To the best of our knowledge, the proposed formalism is general enough to potentially encompass any known learning process providing or not providing explanations. Indeed, the objects $X, Y, Y^*, E$ can be instantiated through a suitable translator functor to have the desired characteristics, e.g. $\mc{T}(Y)$ could include the explanations space, making the explanations optimizable, or $\mc{T}(Y^*) = \{*\}^{\mathbb{N}}$, giving an unsupervised model. Further examples showing the flexibility of our  framework can be found in the Appendix \ref{app:ALA}.
\subsection{Finding \#2: Our framework enables a formal definition of ``explanation''}
\label{sec:syExp}

Our theoretical framework allows us to provide the first formal definition of the term ``explanation'', which embodies the very essence and purpose of explainable AI. Our formalization goes beyond a mere definition of ``explanation'', as it highlights a natural distinction between different forms of explanations. Indeed, institution theory 
allows a straightforward characterization of syntactic and semantic explanations. While both forms of explanations are prevalent in the current XAI literature, their distinctions are often overlooked, thus limiting a deep understanding of the true nature and intrinsic limitations of a given explanation.

\begin{definition} 
    Given an institution $I$, an object $\Sigma$ of $\cat{Sign}_I$, and a concrete explainer $\hat{\eta}=\mc{T}(\eta):\mc{X}\times \mc{P}\rightarrow \mc{Y}\times \mc{E}$, an \emph{explanation} $\mc{E} = \mc{T}(E)$ in a language $\Sigma$ is a set of $\Sigma$-sentences (\emph{syntactic} explanation) or a model of a set of $\Sigma$-sentences (\emph{semantic} explanation). 
\end{definition}

We immediately follow up our definition with a concrete example to make it more tangible.

\begin{example}\label{ex:logsig}
    Let $I_{PL}$ be the institution of Propositional Logic and $\Sigma$ a signature of $I_{PL}$ such that $\{x_{\textit{flies}},x_{\textit{animal}},x_{\textit{plane}},x_{\textit{dark\_color}},\ldots\}\subseteq\Sigma$ and with the standard connectives of Boolean Logic, i.e. $\neg,\wedge,\vee,\rightarrow$. For instance, $\hat{\eta}$ could be an explainer aiming at predicting an output in $\mc{Y}=\{x_{\textit{plane}},x_{\textit{bird}},\ldots\}$ given an input in $\mc{X}$. Then a syntactic explanation could consist of a $\Sigma$-sentence like $\varepsilon = x_{\textit{flies}}\wedge\neg x_{\textit{animal}}\rightarrow x_{\textit{plane}}$, a semantic explanation could be the truth-function of  $\varepsilon$.
\end{example}

Our definition of explanation not only generalizes and formalizes existing definitions in the literature but also encompasses key aspects discussed by different researchers. For instance, according to ~\citet{miller2019explanation}, an explanation is ``an answer to a 'why?' question''; \citet{Das2020OpportunitiesAC} define explanation as ``additional meta information, generated by an external algorithm or by the machine learning model itself, to describe the feature importance or relevance of an input instance towards a particular output classification''; while \citet{palacio2021xai} describe explanation as ``the process of describing one or more facts, such that it facilitates the understanding of aspects related to said facts''. Our definition of explanation incorporates all these notions, as $\Sigma$-sentences and their models can provide any form of statement related to the explaining learning agent and its inputs/outputs. This encompasses additional meta information and feature relevance~\citep{Das2020OpportunitiesAC}, description of facts related to a learning process~\citep{palacio2021xai}, or insights into why a specific output is obtained from a given input~\citep{miller2019explanation}.

Furthermore, \citet{tarski1944semantic} and~\citet{goguen1992institutions} proved how the semantics of ``truth'' is invariant under change of signature. This means that we can safely use signature morphisms to switch from one ``notation'' to another, inducing consistent syntactic changes in a $\Sigma$-sentence without conditioning the ``meaning'' or the ``conclusion'' of the sentence~\citep{goguen1992institutions}. As a result, signature morphisms can translate a certain explanation between different signatures, hence paving the way to study ``communication" as well as ``understanding" between XLAs.  

\subsection{Finding \#3: Our framework provides a theoretical foundation for XAI taxonomies}
\label{sec:tax}

Using  our categorical constructions, we can develop a theory-grounded taxonomy of XAI methods that goes beyond current ones and catalogues existing approaches in a systematic and rigorous manner. We recognize the importance of such a foundation due to the ongoing debates and challenges faced by current taxonomies in comprehensively encompassing all existing XAI methods. Indeed, existing approches in the XAI literature have only provided subjective viewpoints on the field, distinguishing methods according to controversial criteria such as: the scope/methodology/usage of explanations~\citep{Das2020OpportunitiesAC}; the how/what/why of explanations~\citep{palacio2021xai}; the relationship between explainers and systems being explained, e.g., intrinsic/post-hoc, model-specific/agnostic, local/global~\citep{molnar2020interpretable}; or the specific data types, such as images~\citep{kulkarni2022explainable}, graphs~\citep{li2022survey}, natural language~\citep{danilevsky2020survey}, and tabular data~\citep{di2022explainable}. However, these taxonomies lack a solid and grounded motivation and rely primarily on subjective preferences. As a result, they are unable to draw universal conclusions and provide a general understanding of the field. On the contrary, our taxonomy aims to fill this gap by providing a comprehensive classification of XAI methods  grounded in our theoretical framework. As an example, we use the proposed categorical framework to explicitly describe the following macro-categories of XAI methods~\citep{Das2020OpportunitiesAC,molnar2020interpretable,palacio2021xai}, where we  distinguish between an LA instantiated by a translator $\mc{T}$, with $\mc{T}(\eta) = \hat{\mu}$, and an XLA instantiated by $\mc{T}'$. We will refer to the objects of the latter using prime, e.g. $\mc{T}'(Y) = \mathcal{Y}'$.


\paragraph{Post-hoc and Intrinsic.}
XAI surveys currently distinguish between intrinsic and post-hoc explainers. Informally, the key difference is that intrinsic XAI methods evolve model parameters and explanations at the same time, whereas post-hoc methods extract explanations from pre-trained models~\citep{Das2020OpportunitiesAC, molnar2020interpretable}.

\textbf{Post-hoc explainer.}
Given a trained LA model $\hat{\mu}: \mathcal{X} \times \mathcal{P} \rightarrow \mathcal{Y}$, a post-hoc explainer is an XLA explainer instantiated by $\mc{T}'$  such that $\hat{\eta}:\mc{X}'\times\mc{P}'\to\mc{Y}'\times\mc{E}'$, with $\mc{X}' = \mathcal{Y} \times \mathcal{X} \times \mathcal{P}$:
\begin{center}\includegraphics[scale=0.37]{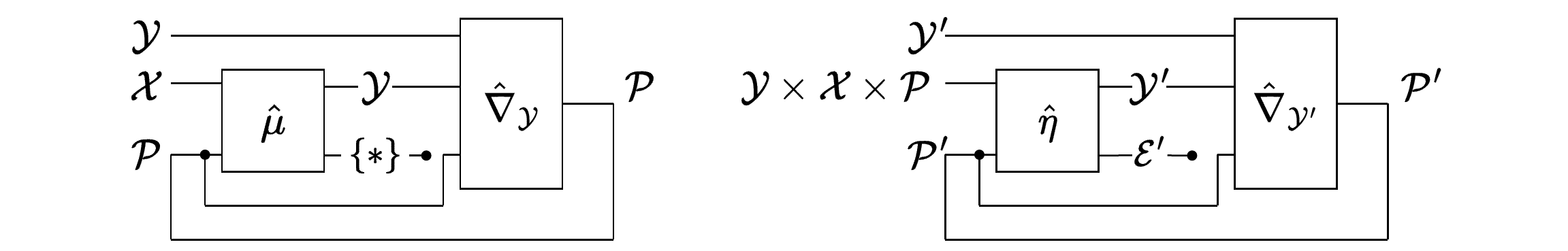}\end{center}

\textbf{Intrinsic explainer.} 
An intrinsic explainer is an XLA explainer $\hat{\eta}$ whose input objects are parameters $\mc{P}'$ and a set of entries of a database $\mathcal{X}'$:
\begin{figure}[!ht]
    \centering
\includegraphics[width=0.20\textwidth]{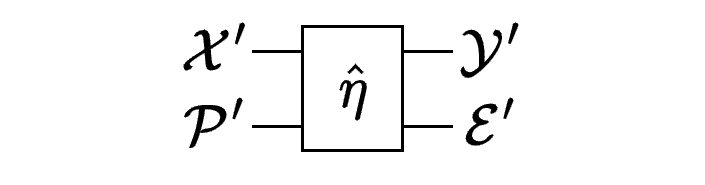}
\end{figure}

Common intrinsic explainers are logic/rule-based~\citep{barbiero2023interpretable, breiman1984classification, ciravegna2023logic, friedman2008predictive, manhaeve2018deepproblog, schmidt2009distilling,yang2017scalable}, linear~\citep{doshi2015graph, hastie2017generalized,nelder1972generalized,santosa1986linear,tibshirani1996regression, verhulst1845resherches}, and prototype-based~\citep{fix1989discriminatory, kaufmann1987clustering} approaches; while well-known post-hoc explainers include saliency maps~\citep{selvaraju2017grad, simonyan2013deep}, surrogate models~\citep{lundberg2017unified,ribeiro2016should,ribeiro2018anchors}, and concept-based approaches~\citep{espinosa2022concept,ghorbani2019interpretation,ghorbani2019towards,kim2016examples}.

\paragraph{Model-Agnostic and Model-Specific}
Intuitively, model-agnostic explainers extract explanations independently from the architecture and/or parameters of the model being explained, whereas model-specific explainers depend on the architecture and/or parameters of the model.

\textbf{Model-agnostic explainer. }
Given an LA model $\hat{\mu}: \mathcal{X} \times \mathcal{P} \rightarrow \mathcal{Y}$, a model-agnostic explainer is an XLA explainer $\hat{\eta}:\mc{X}'\times\mc{P}'\to\mc{Y}'\times\mc{E}'$, such that $\mc{X}' = \mathcal{Y} \times \mathcal{X}$.
\begin{figure}[!h]
\hspace{1cm}
\includegraphics[scale=0.50]{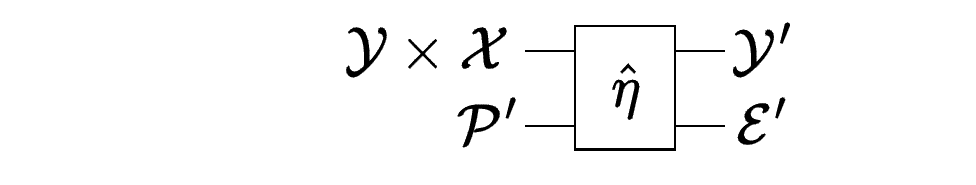}
\end{figure}

\textbf{Model-specific explainer.} 
A model-specific explainer simply differentiates from a model-agnostic, as the XLA explainer $\hat{\eta}:\mc{X}'\times\mc{P}'\to\mc{Y}'\times\mc{E}'$ has $\mc{X}' = \mc{Y}\times \mc{X} \times \mathcal{P}$.
\begin{figure}[!h]
\hspace{1cm}
\includegraphics[scale=0.50]{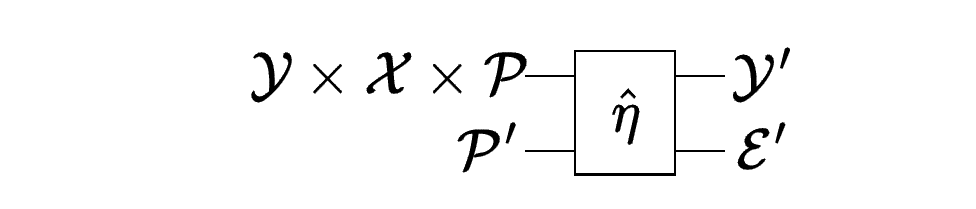}
\end{figure}
Typical examples of model-agnostic explainers include surrogate models~\citep{lundberg2017unified,ribeiro2016should,ribeiro2018anchors} and some concept-based approaches~\citep{ghorbani2019interpretation,ghorbani2019towards,kim2016examples}. Among renowned model-specific explainers instead we can include all model-intrinsic explainers~\citep{molnar2020interpretable} and some post-hoc explainers such as saliency maps~\citep{selvaraju2017grad, simonyan2013deep} as they can only explain gradient-based systems.

\paragraph{Forward and Backward.}
Another relevant difference among XAI methods for gradient-based models is whether the explainer relies on the upcoming parameters~\citep{petsiuk2018rise,zeiler2014visualizing,zintgraf2017visualizing} or the gradient of the loss on the parameters in the learning optimizer~\citep{selvaraju2017grad,simonyan2013deep}.

\textbf{Forward-based explainer. }
Given a gradient-based LA model $\hat{\mu}: \mathcal{X} \times \mathcal{P} \rightarrow \mathcal{Y}$, a forward-based explainer is an XLA explainer $\hat{\eta}:\mc{X}'\times\mc{P}'\to\mc{Y}'\times\mc{E}'$ with $\mc{X}'=\mc{X}''\times \mc{P}$.
\hspace*{1.1cm} 
\includegraphics[scale=0.50]{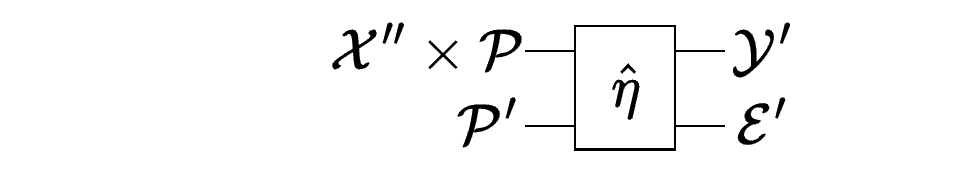}

\textbf{Backward-based explainer. }
Given a gradient-based LA model $\hat{\mu}: \mathcal{X} \times \mathcal{P} \rightarrow \mathcal{Y}$ and an optimizer $\hat{\nabla}_{\mc{Y}}:\mc{Y}\times\mc{Y}\times\mc{P}\rightarrow\mc{P}$, a backward-based explainer is an XLA explainer $\hat{\eta}: \mathcal{X}' \times \mathcal{P}' \rightarrow \mathcal{Y}' \times \mathcal{E}'$ where,  $\mc{X}'=\mc{X}''\times h(\mc{P})$, being $h(\mathcal{P})=\frac{\partial\mathcal{L}(\mc{Y},\mc{Y})}{\partial \mc{P}}$ the gradient of the loss function $\mc{L}$ on $\mc{P}$.

\textbf{A case study: Concept bottleneck models.}
Concept bottleneck models (CBM)~\citep{koh2020concept} are recent XAI architectures which first predict a set of human-understandable objects called ``concepts'' and then use these concepts to solve downstream classification tasks. Our framework allows to formally define even these advanced XAI structures as follows: A CBM is an XLA such that $\hat{\eta}$ is composed of a concept predictor $\hat{\mu}: \mathcal{X} \times \mathcal{P} \rightarrow \mathcal{Y}$ and a task predictor $\hat{\eta}':\mc{Y}\times\mc{P}\to\mc{Y}\times\mc{E}$, $\mathcal{Y}$ is the set of classes and $\mc{P} = \mc{P}' \times \mc{P}''$ is the product of the parameter space of the two models.
\begin{center}\includegraphics[scale=0.50]{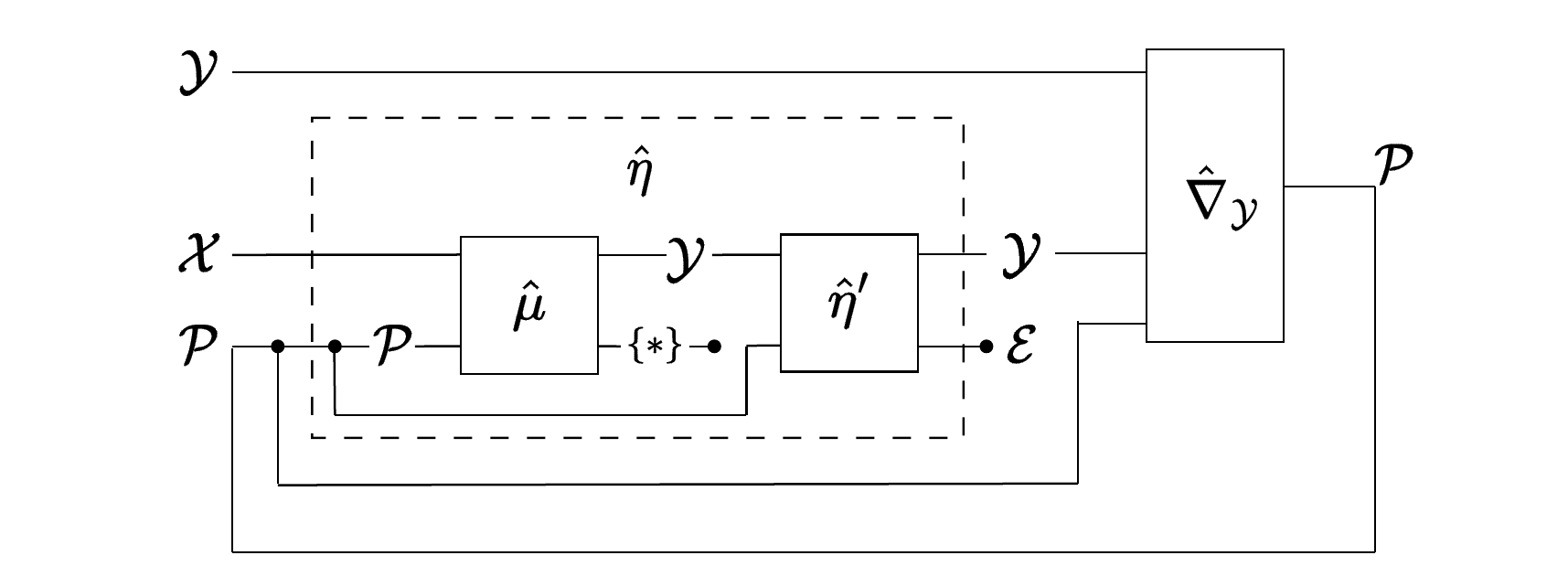}\end{center}

Overall these examples give a taste of the flexibility and expressive power of our categorical framework demonstrating how it can successfully encompass existing XAI approaches and taxonomies.

\subsection{Finding \#4: Our framework emphasizes commonly overlooked aspects of explanations}
\label{sec:xai-semantics}

Current XAI taxonomies often neglect the distinction between syntactic and semantic approaches. On the contrary, our Definition~\ref{def:xla} provides a natural distinction between these two forms of XLAs, thus introducing a novel perspective to analyze XAI methods. On the one hand, syntactic approaches work on the structure of symbolic languages and operate on $\Sigma$-sentences. Notable examples of syntactic approaches are proof systems such as natural deduction~\citep{prawitz2006natural} and sequent calculi~\citep{takeuti2013proof} which are designed to operate on formal languages such as first-order logic.  On the other hand, semantic approaches provide explanations related to the meaning or interpretation of sentences as a model of a language. Most XAI techniques actually fall into this class of methods as semantic explanations establish a direct connection with specific problem domains~\citep{karasmanoglou2022heatmap, lundberg2017unified,ribeiro2016should, simonyan2013deep}. The following examples show the relation between a syntactic and a semantic technique emphasizing connections between XAI methods that often slip unnoticed.

\begin{figure}
\centering
\includegraphics[scale=0.5]{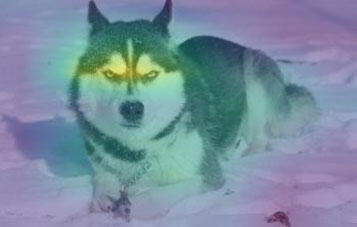}
\caption{Saliency map.}
\label{fig:saliency}
\end{figure}

\begin{example}
    The Gradient-weighted Class Activation Mapping (Grad-CAM),~\citep{simonyan2013deep} is a classic example of a semantic (backward) XLA, whose institution can be defined as a fragment of FOL with all the signatures' objects consisting of a single predicate and a finite set of constants. Intuitively, in a classification task the constants represent the pixels of an image and the relation represents the saliency degree of each pixel for the class prediction. Sentences and models are defined as in FOL by using the provided signature. A general syntactic explanation in this institution can be easily expressed by taking a signature $\Sigma=\{S,p_1,p_2,\ldots\}$ with $S$ the unary saliency predicate and $p_i$ the constant for the $i$-th pixel. Assuming to collect the most ``salient'' pixels in the set $\textit{SalPix}$, the syntactic explanation may be expressed by: $\bigwedge_{p\in\textit{SalPix}}S(p)$. \Cref{fig:saliency} instead represents a semantic explanation where Grad-CAM interprets predicates and constants in the syntactic formula assigning them a meaning (i.e., concrete values).
\end{example}

\begin{figure}
\includegraphics[width=0.40\textwidth]{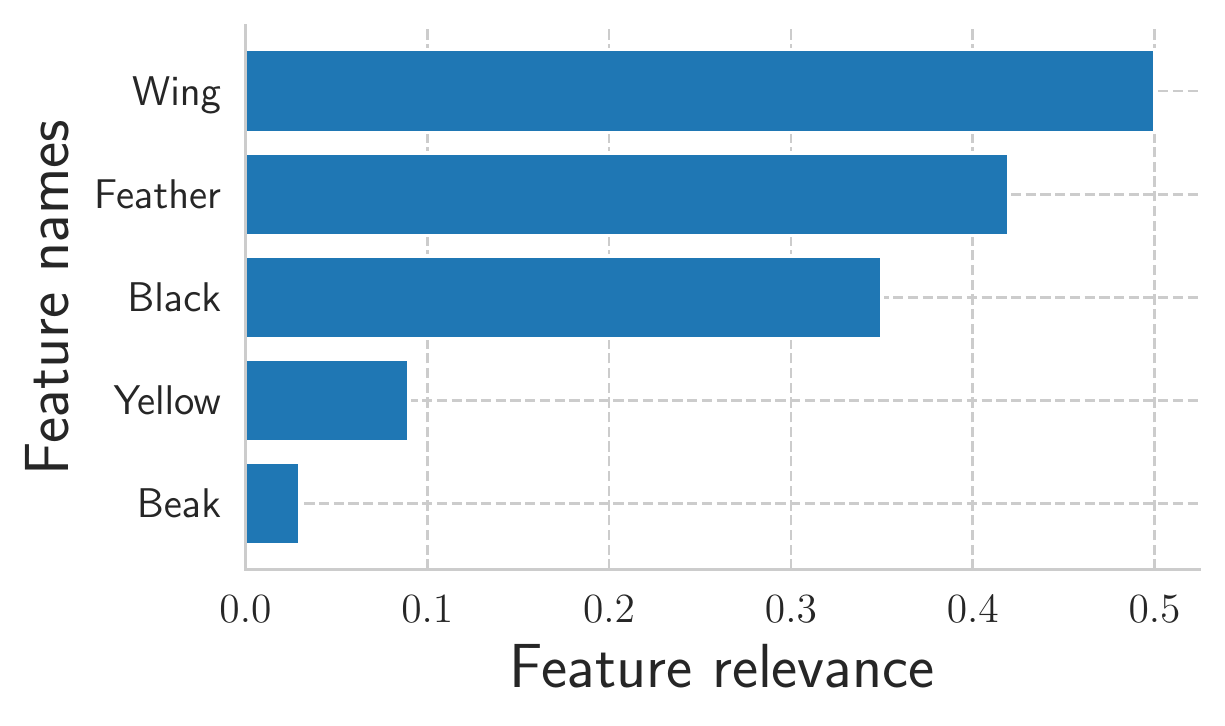}
\caption{Feature importance.}
\label{fig:feat_imp}
\end{figure}

\begin{example} 
Another classic example of semantic XLAs are feature importance methods~\cite{wei2015variable}, such as 
LIME, \cite{ribeiro2016model}. As saliency maps, LIME relies on the institution $I_{un}$, as their signatures consist 
on a set of constants $f_i$ (each for every considered feature) and a single predicate $R$ to express their relevance. Syntactic explanations have the form $\bigwedge_{f\in\textit{ImpFeat}}R(f)$ where the set $\textit{ImpFeat}$ collects the most ``relevant'' features for a task. Figure \ref{fig:feat_imp} shows an example of a semantic explanation of LIME, where the length of each bar represents the relevance of the corresponding feature.
\end{example}

\paragraph{Comparing the expressive power of explanations.}
The Grad-CAM and LIME examples offer the ideal setting to show how our framework can formally assess the expressive power of different forms of explanations, drawing connections between apparently different XLAs. Indeed, provided that both Grad-CAM and LIME signatures contain the same number of pixels/features, these two forms of explanations syntactically possess the same expressive power (while differing in their semantic models and generating algorithms for the explanations). Consequently, we can convert any saliency map into an equivalent feature importance representation and vice versa (using an arity-preserving function), without compromising their meaning/truth value~\citep{tarski1944semantic}. This is generally true, whenever it is possible to define an arity-preserving mapping between different signatures in the same institution. This observation underscores an often overlooked aspect in XAI literature: evaluating the expressive power of explanations requires comparing their signatures and syntax, not the way these explanations are visualized. User studies that compare different forms of visualization essentially assess the visualization's expressive power, which relates to human understanding, rather than the expressive power of an explanation itself.

\paragraph{Limitations of semantic explanations.} The connection between a semantic explanation and a specific context (e.g., an image) may stimulate human imagination, but it often limits the scope and robustness of the explanation, hindering human understanding in the long run~\citep{ghorbani2019interpretation,rudin2019stop}. On the contrary, symbolic languages such as first-order logic or natural language are preferable for conveying meaningful messages as explanations as suggested by~\citet{kahneman2011thinking,marcus2020next}. For this reason, a promising but often overlooked research direction consists in accurately lifting semantic explanations into a symbolic language in order to provide a syntactic explanation~\citep{breiman1984classification, letham2015interpretable,costa2018automatic,guidotti2018local, ribeiro2018anchors,ciravegna2023logic}. By recognizing the importance of this distinction, our formalization can provide a suitable basis to gaining deeper insights into the limitations of different forms of explanations.

\section{Discussion}
\label{sec:key}

\paragraph{Significance and relations with other XAI foundational works.}
The explainable AI research field is growing at a considerable rate~\citep{minh2022explainable} and it now has a concrete impact on other research disciplines~\citep{cranmer2019learning,davies2021advancing,jimenez2020drug} as well as on the deployment of AI technologies~\citep{duran2021afraid,lo2020ethical,wachter2017counterfactual}. This rising interest in explainable AI increases the need for a sound foundation and taxonomy of the field~\citep{adadi2018peeking,palacio2021xai}. Indeed, existing reviews and taxonomies significantly contribute in: (i) describing and clustering the key methods and trends in the XAI literature~\citep{molnar2020interpretable}, (ii) proposing the first qualitative definitions for the core XAI terminology~\citep{Das2020OpportunitiesAC}, (iii) relating XAI methodologies, aims, and terminology with other scientific disciplines~\citep{miller2019explanation}, and (iv) identifying the key knowledge gaps and research opportunities~\citep{Das2020OpportunitiesAC}. However, most of these works acknowledge the need for a more sound mathematical foundation and formalism to support the field. Our framewok arises to fill this gap. In particular our methodology formalizes key XAI notions for the first time, using the category of Cartesian streams and the category of signatures. Our work also draws from~\citet{katsumata19} who propose Cartesian streams to model gradient-based learning, and~\citet{cruttwell2022categorical} who model gradient-based learning using the category of lenses~\citep{riley2018categories}. The categorical formalisms of lenses and streams are closely related~\citep{monoidalStreams}. Intuitively, lenses can be used to encode one-stage processes, while streams can encode processes with time indexed by natural numbers, i.e. providing a more suitable description of the dynamic process of learning. However, our work opens to more general AI systems which are not necessarily gradient-based by generalizing the category of lenses with Cartesian streams.

\paragraph{Limitations}
In this work we face the challenging task of formalizing (previously informal) notions such as ``explanation'', while acknowledging the ongoing debate over their meaning, not only within the AI community but also in philosophy, epistemology, and psychology. Our formalization offers a robust theory-grounded foundation for explainable AI, but it does require readers to engage with abstract categorical structures. However, embracing this initial challenge brings a substantial payoff by enabling us to achieve a comprehensive and unified theory of the field, encompassing all the pertinent instantiations of XAI notions, structures, explanations, and paradigms.

\paragraph{Conclusion}
This work presents the first formal theory of explainable AI. In particular, we formalized key notions and processes that were still lacking a rigorous definition. We then show that our categorical framework enables us to: (i) model existing learning schemes and architectures, (ii) formally define the term ``explanation'', (iii) establish a theoretical basis for XAI taxonomies, and (iv) emphasize commonly overlooked aspects of XAI methods, like the comparison between syntactic and semantics explanations. Through this work, we provide a first answer to the pressing need for a sound foundation and formalism in XAI as advocated by the current literature~\citep{adadi2018peeking,palacio2021xai}. While our taxonomy provides guidance to navigate the field, our formalism strengthens the reputation of explainable AI encouraging a safe and ethical deployment of AI technologies. We think that this work may contribute in paving the way for new research directions in XAI, including the exploration of previously overlooked theoretical side of explainable AI, and the mathematical definition of other foundational XAI notions, like ``understandability'' and ``trustworthiness''.

\bibliography{references}

\newpage
\appendix
\onecolumn
\section{Elements of Category Theory}\label{app:cat}
\subsection{Monoidal Categories}\label{app:mon-cat}

The process interpretation of monoidal categories~\citep{Coecke2017,fritz2020} sees morphisms in monoidal categories as modelling processes with multiple inputs and multiple outputs. Monoidal categories also provide an intuitive syntax for them through string diagrams~\citep{joyal1991geometry}. The coherence theorem for monoidal categories~\citep{maclane78} ensures that string diagrams are a sound and complete syntax for them and thus all coherence equations for monoidal categories correspond to continuous deformations of string diagrams. One of the main advantages of string diagrams is that they make reasoning with equational theories more intuitive.

\begin{definition}[\citet{eilenberg1945general}]
    A \emph{category} \(\cat{C}\) is given by a class of \emph{objects} $\cat{C}^o$ and, for every two objects \(X,Y \in \cat{C}^o\), a set of \emph{morphisms} $\hom(X,Y)$ with input type \(X\) and output type \(Y\). A morphism \(f \in \hom(X,Y)\) is written \(f \colon X \to Y\). For all morphisms \(f \colon X \to Y\) and morphisms \(g \colon Y \to Z\) there is a \emph{composite} morphisms \(f \dcomp g \colon X \to Z\). For each object \(X \in \cat{C}^o\) there is an \emph{identity} morphism \(\id{X} \in \hom(X,X)\), which represents the process that ``does nothing'' to the input and just returns it as it is. Composition needs to be associative, i.e. there is no ambiguity in writing \(f \dcomp g \dcomp h\), and unital, i.e. \(f \dcomp \id{Y} = f = \id{X} \dcomp f\) (\Cref{fig:category-string-diagrams-app}, bottom).
\end{definition}

\begin{figure}[ht!]
    \centering
    $\sequentialFig{}$ \qquad \(\identityXFig\)\\[4pt]
    \compositionUnitalFig{}
    \caption{String diagrams for the composition of two morphisms (top, left), the identity morphism (top, right), and the unitality condition (bottom).}\label{fig:category-string-diagrams-app}
\end{figure}

Monoidal categories~\citep{maclane78} are categories endowed with extra structure, a monoidal product and a monoidal unit, that allows morphisms to be composed \emph{in parallel}. The monoidal product is a functor \(\tensor \colon \cat{C} \times \cat{C} \to \cat{C}\) that associates to two processes, \(f_1 \colon X_1 \to Y_1\) and \(f_2 \colon X_2 \to Y_2\), their parallel composition \(f_1 \tensor f_2 \colon X_1 \tensor X_2 \to Y_1 \tensor Y_2\) (\Cref{fig:string-diagrams-monoidal-cat-app}, center). The monoidal unit is an object \(e \in \cat{C}^o\), which represents the ``absence of inputs or outputs'' and needs to satisfy \(X \tensor e \iso X \iso e \tensor X\), for each \(X\in \cat{C}^o\). For this reason, this object is often not drawn in string diagrams and a morphism \(s \colon e \to Y\), or \(t \colon X \to e\), is represented as a box with no inputs, or no outputs.
\[\statemorphismFig{} \quad \text{and} \quad \costatemorphismFig{}\]

\subsection{Cartesian and Symmetric Monoidal Categories}
\label{app:symm-mon-cat}

A symmetric monoidal structure on a category is required to satisfy some coherence conditions \citep{maclane78}, which ensure that string diagrams are a sound and complete syntax for symmetric monoidal categories \citep{joyal1991geometry}. 
Like functors are mappings between categories that preserve their structure, \emph{symmetric monoidal functors} are mappings between symmetric monoidal categories that preserve the structure and axioms of symmetric monoidal categories.

A monoidal category is \emph{symmetric} if there is a morphism \(\swap{X,Y} \colon X \tensor Y \to Y \tensor X\), for any two objects \(X\) and \(Y\), called the \emph{symmetry}.
\begin{figure}[h!]
    \centering
    $\morphismManyWiresFig{} \qquad \parallelFig{} \qquad \swapXYFig{X}{Y}$
    \caption{A morphism with multiple inputs and outputs (left), the parallel composition of two morphisms (center), and the symmetry (right) in a monoidal category.}
    \label{fig:string-diagrams-monoidal-cat-app}
\end{figure}
The inverse of \(\swap{X,Y}\) is \(\swap{Y,X}\):
\[\swapinverseFig{X}{Y}\]
These morphisms need to be coherent with the monoidal structure, e.g.
\[\hexagoneqexampleFig{X}{Y}{Z}.\]
Moreover, the symmetries are a natural transformation, i.e. morphisms can be swapped,
\[\swapnaturalFig{}.\]

Some symmetric monoidal categories have additional structure that allows resources to be copied and discarded~\citep{fox76}.
These are called \emph{Cartesian categories}.
It is customary to indicate with \(\times\) the monoidal product given by the cartesian structure and with \(e\) the corresponding monoidal unit.
Cartesian categories are equipped, for every object \(X\), with morphisms \(\cp_X \colon X \to X \times X\) and \(\discard_X \colon X \to e\):
\[\copyXFig{} \quad \text{and} \quad \discardXFig{}\]
These need to be natural transformations, i.e. morphisms can be copied and discarded,
\begin{align*}
    \copynaturalFig{}\\
    \discardnaturalFig{}
\end{align*}
be coherent with the monoidal structure
\begin{align*}
    \copycoherentFig{X}{Y}\\
    \discardcoherentFig{X}{Y}
\end{align*}
and satisfy the equations of a cocommutative comonoid.
\begin{align*}
    \copycoassociativeFig\\
    \copycounitalFig\\
    \copycocommutativeFig
\end{align*}

\subsection{Feedback Monoidal Categories}
\label{app:feedback-cat}

\begin{definition}\label{def:fmc}
A \emph{feedback monoidal category} is a symmetric monoidal category \(\cat{C}\) endowed with an endofunctor $F:\cat{C} \rightarrow \cat{C}$, and an operation \(\fbk[S] \colon \hom (X \times F(S), Y \times S) \to \hom (X,Y)\) for all objects \(X,Y,S\) in \(\cat{C}\),
which satisfies the following axioms:
\[
\small
\begin{array}{ll}
    \mbox{(Tightening)} & \mbox{\(\fbk[S]((g \times \id{FS}) \dcomp f \dcomp (h \times \id{S})) = g \dcomp \fbk[S](f) \dcomp h\);} \\
     \mbox{(Joining)} & \mbox{\(\fbk[S \times T](f) = \fbk[S](\fbk[T](f))\);} \\
     \mbox{(Vanishing)} & \mbox{\(\fbk[\monoidalunit](f) = f\);} \\
     \mbox{(Strength)} & \mbox{\(\fbk[S](g \times f) = g \times \fbk[S](f)\);} \\
     \mbox{(Sliding)} & \mbox{\(\fbk[T](f \dcomp (\id{Y} \times g)) = \fbk[S]((\id{X} \times g) \dcomp f)\)} .
\end{array}
\]
\end{definition}
\emph{Feedback monoidal functors} are mappings between feedback monoidal categories that preserve the structure and axioms of feedback monoidal categories.

Feedback monoidal categories are the \emph{syntax} for processes with feedback loops.
When the monoidal structure of a feedback monoidal category is cartesian, we call it \emph{feedback cartesian category}.
Their \emph{semantics} can be given by monoidal streams~\citep{monoidalStreams}.
In cartesian categories, these have an explicit description.
We refer to them as cartesian streams, but they have appeared in the literature multiple times under the name of ``stateful morphism sequences''~\citep{katsumata19} and ``causal stream functions''~\citep{uustalu05}.

\subsection{Cartesian Streams}
\label{app:streams}

A \emph{cartesian stream} \(\stream{f} \colon \stream{X} \to \stream{Y}\), with \(\stream{X} = (X_0, X_1, \dots)\) and \(\stream{Y} = (Y_0, Y_1, \dots)\), is a family of functions \(f_n \colon X_n \times \cdots \times X_0 \to Y_n\) indexed by natural numbers.
Cartesian streams form a category \(\Stream{\Set}\).

Each \(f_n\) gives the output of the stream at time \(n\).
We can compute the outputs until time \(n\) by combining \(f_0, \dots, f_n\) to get \(\hat{f}_n \colon X_n \times \cdots \times X_0 \to Y_n \times \cdots \times Y_0\) as follows:
\begin{itemize}
\item \(\hat{f}_0 \defn f_0\)
\item \(\hat{f}_{n+1} \defn (\id{X_{n+1}} \times \cp_{X_n \times \cdots \times X_0}) \dcomp (f_{n+1} \times \hat{f}_n)\)
\end{itemize}
\[\cartesianStreamsCompositionFig\]
The composition of two cartesian streams \(\stream{f} \colon \stream{X} \to \stream{Y}\) and \(\stream{g} \colon \stream{Y} \to \stream{Z}\) has components \((\stream{f} \dcomp \stream{g})_n \defn \hat{f}_n \dcomp g_n\), and the identity stream has projections as components \((\id{\stream{X}})_n \defn \proj{X_n}\).

\subsection{Free Categories}
\label{app:free}

We generate ``abstract'' categories using the notion of \emph{free category}~\citep{maclane78}. Intuitively, a free category serves as a template for a  class of categories (e.g., feedback monoidals). To generate a free category, we just need to specify a set of objects and morphisms generators. Then we can realize ``concrete'' instances of a free category $\cat{F}$ using a functor from $\cat{F}$ to another category $\cat{C}$ that preserves the axioms of $\cat{F}$. If such a functor exists then $\cat{C}$ is of the same type of $\cat{F}$ (e.g., the image of a free feedback monoidal category via a feedback functor is a feedback monoidal category).

\subsection{Institutions}
\label{app:institutions}

An \emph{institution} $I$ is constituted by:
\begin{itemize}
\item[] (i) a category
\(\cat{Sign}_I\) whose objects are signatures (i.e. vocabularies of symbols);
\item[] (ii) a functor $Sen: \cat{Sign}_I \mapsto \cat{Set}$ providing sets of well-formed expressions ($\Sigma$-sentences) for each signature \(\Sigma\in\cat{Sign}_I^o\);
\item[] (iii) a functor $Mod: \cat{Sign}_I^{op} \mapsto \cat{Set}$ providing semantic interpretations, i.e. worlds.
\end{itemize}
Furthermore, Satisfaction is then a parametrized relation $\models_{\Sigma}$ between $Mod(\Sigma)$ and $Sen(\Sigma)$, such that for all signature morphism $\rho: \Sigma \mapsto \Sigma'$, $\Sigma'$-model $M'$, and any $\Sigma$-sentence $e$, 
\begin{equation*}
    M' \models_{\Sigma} \rho(e) \text{ iff } \rho(M') \models_{\Sigma} e
\end{equation*}
where $\rho(e)$ abbreviates $Sen(\rho)(e)$ and $\rho(M')$ stands for $Mod(\rho)(e)$.

\subsection{Expressive power of abstract learning agents}
\label{app:ALA}
Abstract explaining learning agents have the capability of encode a wide range of known learning processes given a suitable translator functor. Indeed, Definition \ref{def:xla} allows to encode any learning scheme with an input ($X$), an optimized output ($Y$), a supervision ($Y^*$), and an unoptimized output ($E$) of any conceivable type. It is important to note that there is no restriction in how $X, Y, Y^*, E$ can be instantiated by a translator functor since, as object in a free category, they act as empty boxes concretely filled by the given translator. The following Tables \ref{tab:lear-sch} and \ref{tab:lear-models} we show the fundamental characteristics of translator functors instantiating various type of learning processes with or without explanations as output of the model. We will use the symbol $\diamond$ as wildcard of dataspace meaning any type of data and $\star$ as a wildcard for the type of functions and we denote with $\mc{T}'$ an additional generic LA.  

\begin{table}
    \centering
    \begin{tabular}{ |p{3.8cm}||p{1cm}|p{1cm}|p{1cm}|p{1cm}|p{1cm}| }
 \hline
 Learning schemes& $\mc{T}(X)$ & $\mc{T}(Y)$ &  $\mc{T}(Y^*)$ & $\mc{T}(E)$ & $\mc{T}(\eta)$\\
 \hline
 Gen. unsup. model   & $\diamond^{\mathbb{N}}$   & $\diamond^{\mathbb{N}}$ & $\{*\}^{\mathbb{N}}$ & $\{*\}^{\mathbb{N}}$ & $\star$\\
 Gen. sup. model   & $\diamond^{\mathbb{N}}$   & $\diamond^{\mathbb{N}}$ & $\diamond^{\mathbb{N}}$ & $\{*\}^{\mathbb{N}}$ & $\star$\\
 Gen. continual lear. model   & $\diamond^{\mathbb{N}}$   & $\diamond^{\mathbb{N}}$ & $\diamond^{\mathbb{N}}$ & $\{*\}^{\mathbb{N}}$ & $\star$\\
 Gen. explaining model   & $\diamond^{\mathbb{N}}$   & $\diamond^{\mathbb{N}}$ & $\diamond^{\mathbb{N}}$ & $\diamond^{\mathbb{N}}$  & $\star$\\
\hline
\end{tabular}
\caption{General learning schemes. In the table the input, output are supposed to be arbitrary but of fixed type (i.e. $\diamond^{\mathbb{N}}$). The architecture is supposed to be potentially variable over the time (i.e. $\star$). To express a fixed architecture with fixed domain is sufficient to add the constraint $\mc{T}(\eta)_i = \mc{T}(\eta)_{i+1}$.}
\label{tab:lear-sch}
\end{table}

\begin{table}
\centering
\begin{tabular}{ |p{2.8cm}||p{3.8cm}|p{1.5cm}|p{1cm}|p{1cm}|p{2.6cm}| }
 \hline
 Learning models& $\mc{T}(X)$ & $\mc{T}(Y)$ &  $\mc{T}(Y^*)$ & $\mc{T}(E)$ & $\mc{T}(\eta)$\\
 \hline
 MLPs   & $\diamond^{\mathbb{N}}$   & $\diamond^{\mathbb{N}}$ & $\diamond^{\mathbb{N}}$ & $\{*\}^{\mathbb{N}}$ & $\mc{T}(\eta)_i = \mc{T}(\eta)_{i+1}$ \\
 RNNs   & $X^{\mathbb{N}} \times H^{\mathbb{N}}$   & $Y^{\mathbb{N}} \times H^{\mathbb{N}}$ & $Y^{\mathbb{N}}$ & $\{*\}^{\mathbb{N}}$ & $\mc{T}(\eta)_i = \mc{T}(\eta)_{i+1}$ \\
 nural arch.
search  & $\diamond^{\mathbb{N}} $    & $\diamond^{\mathbb{N}}$ & $\mc{T}(Y)$ & $\{*\}^{\mathbb{N}}$ & $\star$\\
 CBMs   &  $\diamond^{\mathbb{N}}$   & $\diamond^{\mathbb{N}}$ & $\mc{T}(Y)$ & $\diamond^{\mathbb{N}}$ & $(\hat{\mu}\times \id{P});\hat{\eta}'$\\
 Post-hoc exps   & $\mc{T}'(Y) \times \mc{T}'(X) \times \mc{T}'(P) $    & $\diamond^{\mathbb{N}}$ & $\mc{T}(Y)$ & $\diamond^{\mathbb{N}}$ & $\mc{T}(\eta)_i = \mc{T}(\eta)_{i+1}$ \\
 Intrin. exps   & $\diamond^{\mathbb{N}}$ & $\diamond^{\mathbb{N}}$ & $\mc{T}(Y)$ & $\diamond^{\mathbb{N}}$ & $\mc{T}(\eta)_i = \mc{T}(\eta)_{i+1}$ \\
 Model-agn. exps   & $\mc{T}'(Y) \times \mc{T}'(X)$    & $\diamond^{\mathbb{N}}$ & $\mc{T}(Y)$ & $\diamond^{\mathbb{N}}$ & $\mc{T}(\eta)_i = \mc{T}(\eta)_{i+1}$ \\
 Model-spc. exps   & $\mc{T}'(Y) \times \mc{T}'(X) \times \mc{T}'(P) $    & $\diamond^{\mathbb{N}}$ & $\mc{T}(Y)$ & $\diamond^{\mathbb{N}}$ & $\mc{T}(\eta)_i = \mc{T}(\eta)_{i+1}$\\
 Backw.-base exps    & $\diamond^{\mathbb{N}} \times \mc{T}'(P)$    & $\diamond^{\mathbb{N}}$ & $\mc{T}(Y)$ & $\diamond^{\mathbb{N}}$ & $\mc{T}(\eta)_i = \mc{T}(\eta)_{i+1}$ \\
 Forw.-base exps    & $\diamond^{\mathbb{N}} \times h(\mc{T}(P))$    & $\diamond^{\mathbb{N}}$ & $\mc{T}(Y)$ & $\diamond^{\mathbb{N}}$ & $\mc{T}(\eta)_i = \mc{T}(\eta)_{i+1}$ \\
 \hline
\end{tabular}
\caption{Standard learning models. In the table $H$ is the space of the state vector. For the explainers, $\mc{T}'$ is a translator instantiating an LA to be explained, $h(\mathcal{P})=\frac{\partial\mathcal{L}(\mc{Y},\mc{Y})}{\partial \mc{P}}$ is the gradient of the loss function $\mc{L}$ on $\mc{P}$, and $\mc{T}(\eta)_i = \mc{T}(\eta)_{i+1}$ represents a fixed architecture with fixed domain.}
\label{tab:lear-models}
\end{table}

\end{document}